\RequirePackage[l2tabu, orthodox]{nag}

\documentclass[fontsize=12pt,a4paper,numbers=enddot,parskip=half]{scrartcl} 

\usepackage{float}

\usepackage[T1]{fontenc}
\usepackage{lmodern} 

\usepackage{amsfonts}
\usepackage{amsmath}
\usepackage{amsthm}
\usepackage{amssymb}
\usepackage{eurosym}
\usepackage{bbm}

\setkomafont{section}{\rmfamily\fontseries{b}\selectfont}
\setkomafont{subsection}{\rmfamily\itshape\mdseries}
\setkomafont{subsubsection}{\rmfamily\itshape\mdseries}
\setkomafont{paragraph}{\rmfamily\itshape\mdseries}
\setkomafont{title}{\rmfamily\large\fontseries{b}\selectfont}
\setkomafont{captionlabel}{\rmfamily\large\fontseries{b}\small\selectfont}

\usepackage{multirow}
\usepackage[font={small},justification=raggedright,singlelinecheck=false]{caption}
\usepackage[activate={true,nocompatibility},final,tracking=true,kerning=true,spacing=true]{microtype}
\newcommand{\mytablefont}{\vspace{2mm}\fontsize{9}{10.8} \selectfont}

\usepackage{tikz}

\usetikzlibrary{positioning,arrows}

\usepackage[caption=false]{subfig}
\usepackage{todonotes}

\usepackage[ansinew]{inputenc}
\usepackage{csquotes}
\usepackage[english]{babel}

\begin{hyphenrules}{english}
\end{hyphenrules}

\usepackage[top=3cm, bottom=3cm, left=3cm, right=3cm]{geometry}
\usepackage{setspace}
\usepackage{rotating} 
\usepackage{dcolumn}
\newcolumntype{d}[1]{D{.}{.}{#1}}

\usepackage{graphicx}
\usepackage{booktabs}
\usepackage{epstopdf}
\usepackage{enumerate}
\usepackage[hang]{footmisc}
\usepackage[longnamesfirst]{natbib}
\usepackage{authblk}
\usepackage{colortbl}
\usepackage{placeins}
\usepackage{tabularx}
\usepackage{makecell}
\usepackage{wrapfig}
\newcolumntype{Y}{>{\centering\arraybackslash}X}
\newcolumntype{R}{>{\flushright\arraybackslash}X}
\newcolumntype{L}[1]{>{\raggedright\let\newline\\\arraybackslash\hspace{0pt}}m{#1}}

\author{
\vspace{5mm} \small MORITZ SCHERRMANN\thanks{Institute for Finance \& Banking, Ludwig-Maximilians-Universit\"at M\"unchen, Ludwigstr.\ 28 RB, 80539 Munich, Germany. E-mail: scherrmann@lmu.de}}
\title{\Large German FinBERT: A German Pre-trained Language Model 
for Financial Textual Data}
\date{}

\begin{document}

\maketitle
\vspace{2cm}
{\rmfamily\fontseries{b}\selectfont Abstract}

\smallskip
\small \singlespacing This study presents German FinBERT, a novel pre-trained German language model tailored for financial textual data. The model is trained through a comprehensive pre-training process, leveraging a substantial corpus comprising financial reports, ad-hoc announcements and news related to German companies. The corpus size is comparable to the data sets commonly used for training standard BERT models. I evaluate the performance of German FinBERT on downstream tasks, specifically sentiment prediction, topic recognition and question answering against generic German language models. My results demonstrate improved performance on finance-specific data, indicating the efficacy of German FinBERT in capturing domain-specific nuances. The presented findings suggest that German FinBERT holds promise as a valuable tool for financial text analysis, potentially benefiting various applications in the financial domain.
\normalsize
\vspace{2cm}

\newpage
\onehalfspacing

\section{Introduction}
\label{paper3_sec_introduction}
The digital age has ushered in an unprecedented surge in the volume of information available to individuals and institutions. A significant portion of this information is presented in an unstructured format, primarily as textual data. This proliferation is evident in various sources, ranging from traditional outlets such as newspapers and websites to more contemporary platforms like social media. While the immense volume of data poses challenges, it also provides opportunities for those with the appropriate tools to glean valuable insights from this abundance of information.

In the financial area, the ability to harness and interpret this vast amount of textual data is of paramount importance. Capital market practitioners and researchers are increasingly recognizing the potential of Natural Language Processing (NLP) techniques to distill actionable information from financial texts. Such insights can significantly enhance investment decisions and strategies in various ways. For instance, by ascertaining the dominant theme within a corpus of texts, such as news articles over a specified time frame, one can discern emergent market trends or alterations in sectoral emphasis. This can be instrumental in anticipating market movements or understanding emergent areas of interest. Another example is the investigation of the sentiment of texts, like social media posts. Gauging the tonality embedded within texts can provide a window into the collective mindset of investors regarding a particular company or sector. A positive or negative sentiment can be indicative of potential bullish or bearish market movements, respectively. A further valuable information that might be encapsulated in financial texts are named entities, like performance measures or financial indicators. Extracting such entities from unstructured texts allows for a more structured representation of data. This structured data can then be more readily integrated into analytical models or used for comparative analyses.

The BERT model (Bidirectional Encoder Representations from Transformers, \cite{bert}) has emerged as a powerful tool for these tasks. BERT operates by considering both the preceding and following context in a sentence, allowing for a more nuanced understanding of the text. Its architecture, which pre-trains on a vast corpus and fine-tunes on specific tasks, has proven effective in capturing contextual nuances and producing state-of-the-art results across various NLP tasks.

While BERT's generic models have shown considerable capabilities for a wide area of different text-related tasks, recent research has underscored the benefits of domain-specific pre-training. Models pre-trained on domain-specific corpora tend to outperform generic models when applied to tasks within that domain. Prominent instances within the financial sector encompass FinBERT \citep{finbert}, designed for English-language financial texts, and SecBERT \citep{secbert}, specialized for entity extraction from annual and quarterly reports of US-based corporations. These models underscore the advantages of domain-specific pre-training, as they capture the unique lexicon and semantic structures inherent to the financial domain.

However, a notable gap exists in the realm of domain-specific models: the linguistic diversity of financial markets. While models like FinBERT cater to English texts, the global nature of finance necessitates models tailored to other languages. Recognizing the significance of the German financial market and the wealth of German-language financial texts, this research introduces the German FinBERT. This model is designed to process and analyze German financial texts, facilitating more nuanced analyses of German companies and the broader German-speaking financial landscape.

The cornerstone of my data set comprises annual reports of German companies, sourced from the German Federal Gazette (Bundesanzeiger). These reports, being comprehensive and rich in financial lexicon, serve as a valuable resource for domain-specific pre-training. Supplementing this primary data set are other German textual sources, including news articles, ad-hoc announcements and relevant Wikipedia articles. These additional sources ensure a more holistic coverage of the financial domain, capturing both the formal and informal nuances of the sector.

Given the extensive length of some of these documents, especially annual reports, I employ an 	inevitable data preparation process. To ensure the retention of as much valuable information as possible, I segment long documents, minimizing data loss due to truncation to the maximum sequence length of 512 of standard BERT models. This segmentation ensures that the model is exposed to a diverse range of financial contexts and terminologies.

I train the German FinBERT using a standard BERT pre-training approach. However, in a departure from the traditional BERT methodology and following more recent research, I omit the next sentence prediction task. I explore two distinct training methodologies in this research. The first approach is the training from scratch. This approach involves training the BERT model with random initialized weights, without any prior knowledge, solely on the German financial texts. The second approach is further pre-training. Here, a generic German BERT model, already pre-trained on a broad corpus, is further fine-tuned using the domain-specific financial data.

Upon evaluation, the further pre-trained version of the German FinBERT model demonstrates superior performance compared to generic German BERT models, particularly in finance-specific downstream tasks. This superiority underscores the value of domain-specific pre-training, especially in specialized sectors like finance where the lexicon and contextual nuances can significantly differ from general language. For the pre-trained from scratch version of the German FinBERT model, the results are mixed. Nevertheless, this iteration exhibits signs of undertraining, necessitating further tests by adding more pre-training steps.

The following sections will provide a comprehensive review of the recent literature, elaborate on the processes involved in the creation and preparation of the training corpus, elucidate the BERT pre-training setup and present empirical evaluations that validate the findings discussed herein.

\FloatBarrier
\section{Literature Review}
\label{paper3_sec_literature}
 BERT is a transformer-based \citep{transformer} deep learning model for natural language processing (NLP). It utilizes a bidirectional approach to encode word representations, enabling it to capture contextual information for every word from both left and right side in a text. In the original paper, BERT's pre-training involves masked language modeling (MLM) and next sentence prediction (NSP), both unsupervised tasks that allow the usage of vast amounts of unlabeled text data. In MLM, certain tokens within a sentence are randomly masked and the model is trained to predict these masked tokens based on their surrounding context. NSP, on the other hand, involves training the model to discern whether a given sentence logically follows a preceding sentence, thereby enhancing the model's understanding of sentence relationships. Fine-tuning BERT on downstream tasks has demonstrated its ability to achieve state-of-the-art performance across various natural language understanding tasks, making it a widely adopted and influential model in the field of NLP.

Originally, the pre-training of BERT is designed to make it a versatile and general model, making it applicable to a wide range of different tasks of different domains. The authors achieve this property by pre-training BERT on general domain corpora, such as news articles, books or Wikipedia. The performance of such a general model is typically measured and compared by its performance on the GLUE benchmark, which is a collection of diverse natural language understanding tasks designed to evaluate the performance of language models across a range of linguistic capabilities \citep{wang2018glue}. While its high GLUE score suggests that the original BERT model excels across a wide range of tasks, research has shown that pre-training BERT models on large-scale domain-specific corpora can further enhance their performance on domain-specific tasks compared to a generic model.

For instance, BioBERT \citep{biobert} is a BERT model that is pre-trained on biomedical text, pre-trained on more than 21 billion words. ClinicalBERT \citep{clinicalbert} is tailored for clinical notes and hospital readmission, similar to the work of \cite{alsentzer2019publicly}. Furthermore, SciBERT targets scientific publications gathered from semantic scholar (3.17 billion tokens) to improve performance in downstream tasks that are related to scientific texts. LegalBERT \citep{legalbert} is a family of BERT models for the legal domain intended to assist legal NLP research, computational law and legal technology applications. It is pre-trained on more than 350.000 legal documents from several fields (e.g., legislation, court cases or contracts).\\
Also in the context of the financial domain, there have been previous publications introducing BERT models that were purposefully trained and tailored for financial texts. \cite{finbert} published FinBERT, a BERT Model for financial communications, pre-trained on a large financial communication corpus of 4.9 billion tokens including corporate reports, earnings conference call transcripts and analyst reports. There is also a multilingual version of FinBERT which is able to process documents of German language among others \citep{desola2019finbert}. However, several studies found that multilingual models were outperformed by single language models \citep{gbert, martin2019camembert}. \cite{secbert} focus their work on a BERT model, called SecBERT, which is able to do XBRL tagging, a named entity recognition (NER) task for financial figures. This task differs from usual NER tasks by the fact that the number of different entities is higher than usual and most financial figures are numeric. Especially, the latter is a problem for BERT-based models as the tokenization of numbers harms the performance. Therefore, the authors experiment with two approaches to replace numeric tokens with pseudo-tokens, which significantly boosts the performance of their BERT models in the specific NER task. They train their models on approximately 10,000 annual and quarterly English reports of publicly traded companies, resulting in a corpus of more than 56 million tokens. 

All the referenced studies consistently demonstrate that domain-specific models yield superior results in domain-specific tasks compared to generic BERT models. In addition, some of these studies encompass investigations into other pertinent aspects. \cite{legalbert} state that there are in general three strategies when applying BERT in specialised domains. These are: (a) Usage of a generic BERT model out of the box, (b) adapt a generic BERT model by additional pre-training on domain-specific corpora and (c) pre-train BERT from scratch on domain-specific data. They find further pre-training and pre-training from scratch both outperform the usage of generic BERT models. In their legal context, both strategies perform equally well. This would imply that further pre-training should be preferred since this approach is less data-intensive and hence less expensive. However, I will test both approaches in this study as my data set is larger than the one used for pre-training LegalBERT and both data sets differ significantly with respect to their structure and vocabulary. \\
Additionally, \cite{legalbert} found that the recommended hyper-parameter range of \cite{bert} for fine-tuning BERT on downstream tasks is suboptimal, which is why they recommend a broader hyper-parameter range. I will follow their suggestions in this study. In addition, several studies have examined the application of tokenizers tailored to the specific corpora utilized for pre-training, recognizing that the lexicon of domain-specific texts may diverge considerably from that of generic textual content \citep{finbert,scibert}. These studies find that using domain-specific tokenizers improve the results, but only by a small margin. Consequently, this study does not employ a specialized tokenizer.\\
\cite{dai2020cost} released two pre-trained BERT models trained on tweets and forum text. But more importantly, they state that often the pre-training data used in domain-specific BERT-models is selected based on subject matter rather than objective criteria. Therefore, they investigate the correlation of similarity between training and target data and task accuracy using different domain-specific BERT models. They find that simple similarity measures can be used to nominate in-domain pre-training data. \\
As gathering domain-specific textual data is a costly and difficult process, \cite{sanchez2022effects} conduct a series of experiments by pre-training BERT models with different sizes of biomedical corpora. Their results demonstrate that pre-training on a relatively small amount of in-domain data with limited training steps can lead to a better performance on downstream domain-specific NLP tasks compared to fine-tuning models pre-trained on general corpora. This motivates the training of domain-specific BERT models even if the database is sparse. However, they find that adding more domain-specific data further improves the results.

\section{Financial Corpora}
\label{paper3_sec_corpus}
\subsection{Pre-training Data}
\label{paper3_sec_pre-training_data}
\begin{table}
\begin{flushleft}
\caption{German FinBERT Pre-train Corpus}
 \mytablefont{This table presents statistics of all data sources of the final corpus used for pre-training of the German FinBERT model. I count the number of documents, sentences, words and tokens for every data source separately. Additionally, I report the data source sizes in gigabytes (GB). I use \textit{spaCy} for sentence segmentation. I tokenize documents using the WordPiece tokenizer of the German \textit{bert-base-german-cased} model. (T = Thousand, M = Million, B = Billion)}
\label{paper3_corpus_datails}
\mytablefont
\begin{tabularx}{\textwidth}{L{2.4cm} *{6}{Y}}
\toprule
Source & Period & Num. Documents & Num. Sentences & Num. Words & Num. Tokens & Size (GB) \\
\midrule
Bundesanzeiger & 2006 - 2023 & 11.69 M & 367.26 M & 7.30 B & 9.88 B & 52.06 \\[0.3cm]
Handelsblatt & 2001 - 2023 & 128.79 T & 3.18 M & 55.57 M & 71.68 M & 0.36 \\[0.3cm]
MarketScreener & 2022 - 2023 & 107.58 T & 1.89 M & 50.15 M & 67.31 M & 0.29 \\[0.3cm]
FAZ & 2014 - 2021 & 47.22 T & 1.24 M & 23.56 M & 30.20 M & 0.15 \\[0.3cm]
Ad-Hoc Announcements & 1996 - 2022 & 65.69 T & 803.07 T & 18.64 M & 24.97 M & 0.12 \\[0.3cm]
LexisNexis \& Event Registry & 2003 - 2023 & 37.35 T & 814.07 T & 17.78 M & 23.64 M & 0.11 \\[0.3cm]
Zeit Online & 1998 - 2023 & 20.43 T & 526.88 T & 9.61 M & 12.27 M & 0.06 \\[0.3cm]
Wikipedia & - & 35.07 T & 158.25 T & 3.61 M & 4.92 M & 0.02 \\[0.3cm]
Gabler Wirtschaftslexikon & 2009 - 2023 & 14.04 T & 128.50 T & 2.83 M & 3.87 M & 0.02 \\
\midrule
Total & 1996 - 2023 & 12.15 M & 376.00 M & 7.48 B & 10.12 B & 53.19 \\
\bottomrule
\end{tabularx}
\end{flushleft}
\end{table}

I pre-train BERT on a large, unlabeled corpus of German data from the financial domain. The data stems from a mix of both, publicly available and private resources. Table \ref{paper3_corpus_datails} provides an overview of all used data sources. The by far largest portion of the data with 11.69 million reports originates from the German Federal Gazette (Bundesanzeiger)\footnote{https://www.bundesanzeiger.de}. The German Federal Gazette is an official publication and announcement medium of the German federal authorities. It is published by the Federal Ministry of Justice with a scope similar to that of the Federal Register in the United States. The Federal Gazette is a mandatory publication journal for judicial and other announcements, for all commercial register entries, as well as for legally required publications of annual financial statements and company deposit notifications. For this study, I only focus on annual financial statements. Annual financial statements provide a consolidated summary of a company's financial transactions and performance, encompassing revenues, expenses, assets, liabilities and equity. They serve as a comprehensive snapshot that enables to assess the company's financial health, resource allocation and strategic direction. In Germany, according to Section 325 of the German Commercial Code, companies including limited liability companies, joint-stock companies, partnership limited by shares and entrepreneurial companies are obligated to disclose information. The German Federal Gazette provides free access to the annual financial statements. I collect the data using a web crawler. The financial reports from the German Federal Gazette are released between 2006 and 2023 and they add up to almost 10 billion tokens, which is more than 52 GB of data. This represents almost 98\% of the final pre-training corpus. 

Another data type I use for the BERT pre-training are news articles. The articles are sourced from newspapers such as \textit{Handelsblatt}\footnote{https://www.handelsblatt.com}, \textit{Frankfurter Allgemeine Zeitung (FAZ)}\footnote{https://www.faz.net} and \textit{Die Zeit}\footnote{https://www.zeit.de}. All the referenced newspapers maintain digital platforms where articles are freely accessible. This facilitates data collection through web scraping methodologies. Additionally, I collect news articles from the \textit{MarketScreener}\footnote{https://de.marketscreener.com}, a financial website with global financial news. I collect further financial news articles from the commercial databases \textit{LexisNexis}\footnote{https://www.lexisnexis.com} and \textit{Event Registry}\footnote{https://eventregistry.org/}.\\
As the focus of this study is to train a finance specific model, it is crucial to ensure that the used textual data stems from the financial domain. As newspapers like the \textit{FAZ} and \textit{Zeit} are general newspapers, an additional filter for business news is necessary. I achieve this by applying an already fitted logistic regression model to distinguish between financial and non-financial news. This is not an issue for news from the Handelsblatt and the MarketScreener as these sources only contain business news. LexisNexis and Event Registry allow to filter for finance related news.\\
All news articles taken together were published between 1998 and 2023 adding up to 205 million tokens (1 GB of data).

I add further types of German financial texts to the corpus like ad-hoc announcements, Wikipedia articles and entries from the \textit{Gabler Wirtschaftslexikon}\footnote{https://wirtschaftslexikon.gabler.de}. I gather the ad-hoc announcements from the EQS Group (formerly Deutsche Gesellschaft f\"ur Ad-hoc-Publizit\"at mbH (DGAP)) \footnote{https://www.eqs-news.com, formerly https://www.dgap.de} between 1996 and 2022.  The Wikipedia articles are filtered for finance related items. The \textit{Gabler Wirtschaftslexikon} is a comprehensive German-language encyclopedia focused on economics and business management topics. However, these data sources constitute only a small portion of the final pre-training corpus.

For all the documents under consideration, the emphasis is placed on textual content, necessitating the exclusion of tables, figures and images.

In total, the final corpus for pre-training consists of more than 12 million German financial texts between 1996 and 2023, which sums up to 10.12 billion tokens, 7.48 billion words and 53.19 GB of data. For comparison: The original BERT model was trained on a corpus size of 3.3 billion words. Similarly, all the mentioned domain-specific BERT models were trained on significantly smaller data sets, with the exception of BioBERT \citep{biobert}, where the data set is comparably large (13.5 billion tokens).

\subsection{Fine-tuning Data}
\label{paper3_sec_finetuning_data}
After BERT has been pre-trained, the MLM performance of the model is not a suitable quality criterion. Since BERT is mainly used for fine-tuning on downstream tasks, the quality of the model should be tested accordingly on such tasks. Unfortunately, there are only a limited number of labeled data sets comprising German finance specific texts that can be used to assess the performance of the BERT model. To the best of my knowledge, the ad-hoc multi-label database of \cite{AdHocMultilabel} is the only manually-labeled, finance-related German textual data set. It consists of 31,771 sentences from 3,044 German ad-hoc announcements. Each sentence is manually allocated to a subset of 20 possible topics.\footnote{In this study, ad-hoc announcements are utilized for both the pre-training and fine-tuning phases of the German FinBERT model. To mitigate the risk of look-ahead bias, all ad-hoc announcements corresponding to the test set of the multi-label ad-hoc database are excluded from the pre-training corpus.} 

In order to test the performance of the model on a broader range of tasks, I create two machine-generated, finance-related German data sets. The first data set is tailored for extractive question answering problems, meaning that the answer to a given question and context is a substring of that context, similar to the SQuAD data set \citep{SQUAD}. The data set is based on the 3,044 announcements of the ad-hoc multi-label database. In a first step, I generate three questions for every document using ChatGPT \citep{ChatGPT}, a state-of-the-art language model developed by OpenAI. To avoid similar questions, I instruct the model to address different aspects of the context with each question. Furthermore, due to the extractive nature of the data set, I instruct the model to only ask questions for which the answer can be given as a substring of the context. This approach results in 9,132 unique context-question pairs. Subsequently, I direct ChatGPT to respond to all queries, emphasizing that its answers must be substrings derived directly from the provided context. Unfortunately, in some cases ChatGPT still produces answers that are not contained in the context, such that only 7,407 observations remain in the data set. See the appendix for further details on the generation of the ad-hoc QuAD database.

The second machine-generated data set is a German version of the financial phrase-bank of \cite{FinSentData1}, which is a sentiment classification data set consisting of manually labeled sentences from news about Finnish companies listed on the OMX Helsinki. Each sentence is allocated to one of three sentiment labels, i.e.~positive, neutral and negative. The data set consists of different splits, depending on the agreement level of the annotators. In this study, I work with the split of the financial phrase bank that contains sentences with at least 75\% agreement between the annotators. After duplicate removal, I get 3,453 labeled sentences. As the original data is given in English, I use the translation tool \textit{DeepL}\footnote{https://www.deepl.com/translator} to transform the data to German.

In addition to the finance-related fine-tuning data sets, this study integrates three German labeled data sets from a general domain. These data sets exhibit similarities with the financial data sets concerning the nature of the problem they address. The primary rationale for incorporating such generic data sets is to assess potential performance degradation of the finance-oriented model when applied to non-specialized data. This methodology offers insights into the model's robustness against variations in the data source. While the model's primary objective is to excel in financial contexts, multifaceted texts, such as news articles, often encompass diverse topics. Consequently, it is imperative for the model to maintain consistent performance across domains beyond finance.

The first generic data set I use in this study is the 10k German News Articles data set (10kGNAD)\footnote{https://github.com/tblock/10kGNAD}. It is based on the One Million Posts Corpus of \cite{schabus2017one}. The data set contains 10,273 news articles of the
Austrian newspaper \textit{Der Standard}. All news on the website of this newspaper are categorized in exactly one of 9 categories, making it a multi-class classification problem. Nevertheless, the task of classifying news to topics loosely relates to the ad-hoc multi-label database.

The next generic data set used in this study is the GermanQuAD data set of \cite{moller2021germanquad}. It is an extractive Q\&A task, similar to the machine generated ad-hoc Q\&A task. The database is constructed from 3,014 unique German Wikipedia articles and consists of 13,722 triplets of contexts, questions and answers. This implies on average more than four questions per article. 

Finally, I add a generic sentiment classification task to the analysis. The respective data set stems from the GermEval 2018 challenge, as presented by \cite{risch2018fine}. The data set consists of German tweets and is characterized by two subtasks, defined by distinct levels of granularity. The first subtask involves a binary sentiment classification, while the second subtask involves a multi-class classification (four classes). In this study, I focus on the binary subtask, where the texts are classified in tweets with or without offense language. The data set consists of 8,407 observations and it is the generic counterpart of the financial phrase-bank. 

I randomly split all fine-tuning data sets in one larger part for training and two smaller parts for validation (e.g.~hyper-parameter tuning) and out-of-sample testing of the model. To do so, I use a stratified sampling approach to maintain the original label distribution within the data splits.

\subsection{Data Preparation}
\label{paper3_sec_data_prep}
\begin{table}
\begin{flushleft}
\caption{Token Distribution Pre-train Corpus (Before \& After Preparation) \& Downstream Tasks}
 \mytablefont{The following table presents summary statistics of the number of tokens per document for both, the pre-train corpus (before and after preparation) and the combined train sets of all downstream tasks.\\(T = Thousand, M = Million, B = Billion)}
\label{paper3_tab_token_distribution}
\mytablefont
\begin{tabularx}{\textwidth}{m{1.5cm} *{8}{Y}}
\toprule
 & Num. Obs. & Min. & 1\% & 10\% & 50\% & 90\% & 99\% & Max. \\
\midrule
Pre-train Corpus & 12.15 M & 1 & 32 & 178 & 440 & 1.03 T & 8.92 T & 203.94 T \\
Prepared Pre-train Corpus & 41.11 M & 11 & 20 & 41 & 216 & 519 & 570 & 23.02 T \\
Fine-tune Tasks & 57.24 T & 3 & 9 & 18 & 54 & 490 & 1.12 T & 4.72 T \\
\bottomrule
\end{tabularx}
\end{flushleft}
\end{table}

A limitation inherent to the BERT model is its constraint on input sequences to a maximum of 512 tokens. Consequently, texts exceeding this token count necessitate truncation for compatibility with the model. It is thus necessary to assess the quantity of documents from the pre-training corpus subject to such truncation and to determine the resultant proportion of the corpus that would be excluded. Table \ref{paper3_tab_token_distribution} shows, among others, summary statistics for the number of tokens per document of the pre-training corpus. We see that for the pre-train corpus, a document has a median length of 440 tokens. Furthermore, 10\% of the data has more than 1,000 tokens and the maximum document length is 203,940 tokens. Almost 40 \% of the documents in the pre-train corpus have more than 512 tokens and hence would be affected by truncation. This would imply that 53 \% of the 10.12 billion tokens would cancel out after truncation. 

Another problem is the token distribution within the pre-train and fine-tune data sets. If we compare the token distribution of the pre-train corpus with the token distribution of the downstream tasks in Table \ref{paper3_tab_token_distribution}, we observe that the documents within the downstream tasks are clearly shorter. The median document length for the downstream tasks is 54 tokens, compared to the 440 tokens in the pre-train corpus. 99\% of the documents of the pre-train corpus consist at least of 32 tokens, which is close to the median of the fine-tune data sets (54). This pattern might cause problems as the German FinBERT might not be used to process short input texts.

To tackle both problems, the significant data loss due to truncation and the different document length between pre-train and fine-tune data, I conduct the following data preparation steps. First, I segment all documents to sentences using \textit{spaCy}\footnote{https://spacy.io}. Second, I sample a minimum number of tokens, with possible values of 30, 100, 300 and 505, and I concatenate sentences of a document until the minimum number of tokens is exceeded. Furthermore, I discard documents shorter than 11 tokens. After that, the procedure repeats until the last sentence of a document. I do not concatenate sentences of different documents. If the minimum number of tokens has not been reached yet, but there is no sentence of a document left, I store the concatenated sentences up to that point as one observation and I move on with the next document.

In addition to the aforementioned preparation steps, I try to manipulate the data as little as possible. The only further step I do is the removal of English documents from the data set, as they might disturb the model pre-training process.

Employing this data preparation methodology results in a threefold increase in the number of observations within the pre-training corpus amounting to 41.11 million. Notably, while the original pre-training corpus experienced a loss of 53\% of tokens due to truncation, the refined approach reduces this loss to a mere 1\% of the tokens. Furthermore, Table \ref{paper3_tab_token_distribution} indicates that the prepared pre-training corpus contains more smaller documents as the median document length decreased from 440 to 216 and 10\% of the data has not more than 41 tokens. This distribution prepares the model better for the downstream tasks. Note that the average number of tokens in a document is still considerably larger compared to the downstream tasks. However, the goal was to increase the number of smaller documents and not to match the distribution of the downstream tasks perfectly. The main goal of this work is to end up with a model that is able to process all types of German financial texts, so it should also be able to handle all types of input lengths. Finally, as observed in Table \ref{paper3_tab_token_distribution}, the prepared data set exhibits a maximum document length of 23,020 tokens. This might initially appear anomalous, suggesting an unusually extended sentence appended to that document. Occasionally, companies present structured data using bullet points rather than the conventional tabular format, resulting in such data not being excluded. Given that these bullet points do not constitute standard sentences, the sentence segmentation algorithm aggregates the information into a singular, elongated sentence. Nonetheless, this anomaly does not pose a challenge, as documents are truncated post the 512-token mark. It's also worth noting that such instances are infrequent, with 99\% of the documents not exceeding 570 tokens in length.

\section{German FinBERT Training}
I pre-train the German FinBERT model from scratch on the prepared corpus introduced in section \ref{paper3_sec_data_prep}. I use the BERT-base architecture from \cite{bert}, including a hidden
size of 768, an intermediate size of 3072, 12 attention heads and 12 hidden layers as well as the GeLU activation function, and learned positional embeddings. Following the idea of \cite{roberta}, I use a sequence length of 512 tokens during the full pre-training. Shorter input documents are padded using the \textit{[PAD]} token, whereas longer input sequences are truncated. I use the tokenizer of the \textit{bert-base-german-cased} model.

Regarding the pre-training procedure itself, I again follow \cite{roberta} by only using the Masked Language Model task (MLM) during pre-training as they observe a decrease in their model performance when they additionally use the Next Sentence Prediction task (NSP), as proposed by the original BERT paper. I apply a 10\% dropout to both the attention and feed-forward layers of all the 12 transformer blocks. I use an Adam optimizer with decoupled weight decay regularization \citep{loshchilov2017decoupled}, with Adam parameters $\beta_1=0.9$, $\beta_2=0.98$, $\epsilon=1\mathrm{e}-6$ and a weight decay of $1\mathrm{e}-5$. The learning rate has a peak value of $5\mathrm{e}-4$, but it linearly decreases close to 0 by the end of the training duration. The first 6\% of the pre-training steps are used for a learning rate warm-up. I train the model using a Nvidia DGX A100 node consisting of 8 A100 GPUs with 80 GB of memory each. With a batch size of 4096, I train the German FinBERT model for 174,000 steps, summing up to more than 17 epochs. I use the code provided by MosaicML\footnote{https://github.com/mosaicml/examples} \citep{mosaicbert}.

Additionally to the pre-training from scratch, I conduct a further pre-training on an already existing, generic German BERT-model, i.e. the \textit{gbert} model of deepset \citep{gbert}. As the model is already pre-trained, I only train the model for further 10,400 steps. Furthermore, I reduce the maximal learning rate to $1\mathrm{e}-4$. The remaining hyper-parameter settings remain consistent with the pre-training from scratch approach.

\FloatBarrier
\section{Experiments}
\subsection{Setup}
In this section, I evaluate the performance of the German FinBERT models on the downstream tasks of section \ref{paper3_sec_finetuning_data}. As benchmark serve three German BERT models that are trained for general purpose with different pre-training strategies and data sets. The first benchmark model I use in this study is the \textit{bert-base-german-cased} model of Deepset\footnote{https://huggingface.co/bert-base-german-cased}. It was the first ever released German BERT model and it was trained with more than 12 GB of textual data from German Wikipedia, the OpenLegalData dump and news articles. The training procedure was similar to the one of the original BERT model of \cite{bert}.

Later, the same company released an improved German BERT model called \textit{gbert-base} \citep{gbert}, which is the next benchmark I use in this study. It has the same structure and number of parameters as its predecessor, but it is trained on a larger data set with 163 GB of text. The predominant segment of the data set is derived from the German subset of the OSCAR corpus \citep{oscar}, accounting for 145 GB. Additionally, the authors have made subtle modifications to the pre-training procedure, opting to mask entire words rather than individual tokens for the MLM objective. While the authors introduce additional models, they either underperform relative to \textit{gbert-base} or possess a greater number of parameters, rendering them non-comparable to the German FinBERT model presented in this research.

The last benchmark of this study is the \textit{gottbert-base} model of \cite{gottbert}. Again, it has the same size as the German FinBERT model and all benchmarks, but the pre-training approach is different as it is copied from the RoBERTa model of \cite{roberta}. The data used by \cite{gottbert} is the German portion of the OSCAR corpus, too. 

I fine-tune all models on all downstream tasks using the \textit{1cycle policy} of \cite{smith2019super}. I
use the Adam optimization method of \cite{kingma2014adam} with standard parameters. For every model, I run a separate grid search on the respective evaluation set for each task to find the best hyper-parameter setup. I test different values for learning rate, batch size and number of epochs, following the suggestions of \cite{legalbert}. After that, I report the results for all models on the respective test set, using the tuned hyper-parameters. 

For every model and task, I repeat that procedure five times with different shuffled training batches and I report all results as averages of the performance measures of these five models. In that way, I reduce the likelihood of ending with good results by chance. 

\subsection{Ad-Hoc Multi-Label Database}
\begin{table}
\begin{flushleft}
\caption{Performance on Ad-Hoc Multi-Label Database}
 \mytablefont{This table compares the performance on the test split of the ad-hoc multi-label database of the German FinBERT model, trained from scratch (SC) and further pre-trained (FP), with the following benchmarks: \textit{bert-base-german-cased}, \textit{gbert-base} \citep{gbert} \& \textit{gottbert-base} \citep{gottbert}. Every model is trained for five different seeds. I report the mean micro and macro F1 scores \citep{sokolova2009systematic} among seeds for every model, together with the standard deviation. I evaluate the performance on document level. All models are trained on sentence level. Numbers are given in percent.}
\label{paper3_tab_model_performance_adhoc_multilabel}
\mytablefont
\begin{tabularx}{\textwidth}{c *{2}{Y}}
\toprule
Model & F1 (Macro) & F1 (Micro) \\
\midrule
bert-base-german-cased & \shortstack[c]{85.37 \\ (0.40)} & \shortstack[c]{85.16 \\ (0.36)} \\[0.3cm]
gbert-base & \shortstack[c]{85.65 \\ (0.25)} & \shortstack[c]{85.20 \\ (0.26)} \\[0.3cm]
gottbert-base & \shortstack[c]{85.29 \\ (0.15)} & \shortstack[c]{84.78 \\ (0.24)} \\
\midrule 
GermanFinBERT\textsubscript{SC} & \shortstack[c]{85.67 \\ (0.61)} & \shortstack[c]{85.17 \\ (0.41)} \\[0.3cm]
GermanFinBERT\textsubscript{FP} & \shortstack[c]{\textbf{86.08} \\ (0.25)} & \shortstack[c]{\textbf{85.65} \\ (0.24)} \\
\bottomrule
\end{tabularx}
\end{flushleft}
\end{table}

Starting with the results for the finance-specific downstream tasks, Table \ref{paper3_tab_model_performance_adhoc_multilabel} displays the outcomes for the ad-hoc multi-label database. I use both macro and micro F1 scores as defined by \cite{sokolova2009systematic}, in accordance with the rationale of \cite{AdHocMultilabel}. Delving into the results, we see that the F1 scores across all models exhibit only minimal differences, with a disparity of less than 1 percentage point between the highest and lowest performing models. The \textit{bert-base-german-cased} model achieves a macro F1 score of 85.37\% with a standard deviation of 0.40\%. The \textit{gbert-base} model slightly outperforms the former with a macro F1 score of 85.65\% and a relatively low standard deviation of 0.25\%. In contrast, the \textit{gottbert-base} model presents a slightly lower macro F1 score of 85.29\% and a standard deviation of 0.15\%. This indicates that the \textit{gbert-base} model performs best on the ad-hoc multi-label database, improving the best result \cite{AdHocMultilabel} by 0.3 percentage points.

Furthermore, the German FinBERT model, when trained from scratch, yields results comparable to the benchmarks, with a macro F1 score of 85.67\%. Its standard deviation, while being the most pronounced among all models, remains relatively modest at 0.61\%. However, the further pre-trained version of German FinBERT surpassed all models, achieving the highest macro F1 score of 86.08\%, surpassing the best benchmark by more than 0.4 percentage points, with a small standard deviation of 0.25\%.

The results are similar for the micro F1 scores. Note that for all models, the gaps between the micro and macro F1 scores are small, indicating that all models perform well also for infrequent topics, which is in line with the findings of \cite{AdHocMultilabel}.
\subsection{Ad-Hoc QuAD}
\begin{table}
\begin{flushleft}
\caption{Performance on  Ad-Hoc QuAD}
 \mytablefont{This table compares the performance on the test split of ad-hoc quad data set of the German FinBERT model, trained from scratch (SC) and further pre-trained (FP), with the following benchmarks: \textit{bert-base-german-cased}, \textit{gbert-base} \citep{gbert} \& \textit{gottbert-base} \citep{gottbert}. Every model is trained for five different seeds. I report the exact matches (EM) and F1 scores \citep{moller2021germanquad} averaged among seeds for every model, together with the standard deviation. Numbers are given in percent.}
\label{paper3_tab_model_performance_adhoc_quad}
\mytablefont
\begin{tabularx}{\textwidth}{c *{2}{Y}}
\toprule
Model & EM & F1 \\
\midrule
bert-base-german-cased & \shortstack[c]{47.61 \\ (0.79)} & \shortstack[c]{71.56 \\ (0.58)} \\[0.3cm]
gbert-base & \shortstack[c]{49.93 \\ (1.46)} & \shortstack[c]{73.30 \\ (1.04)} \\[0.3cm]
gottbert-base & \shortstack[c]{50.66 \\ (0.33)} & \shortstack[c]{73.90 \\ (0.20)} \\
\midrule 
GermanFinBERT\textsubscript{SC} & \shortstack[c]{50.23 \\ (0.69)} & \shortstack[c]{72.80 \\ (0.58)} \\[0.3cm]
GermanFinBERT\textsubscript{FP} & \shortstack[c]{\textbf{52.50} \\ (0.69)} & \shortstack[c]{\textbf{74.61} \\ (0.22)} \\
\bottomrule
\end{tabularx}
\end{flushleft}
\end{table}

Table \ref{paper3_tab_model_performance_adhoc_quad} displays the performance of both German FinBERT models and the benchmark models on the ad-hoc QuAD task. I use two performance metrics in this analysis. First, the exact matches ratio (EM) is utilized, which quantifies the proportion of instances where the model accurately extracts the true answer from the given context. Second, I use the F1 score, considering overlapping tokens between predicted and true answers. 
The \textit{bert-base-german-cased} model achieves an EM of 47.61\% and an F1 score of 71.56\%. The \textit{gbert-base} model, with an EM of 49.93\% and an F1 score of 73.30\%, demonstrates a slight improvement over the former. The \textit{gottbert-base} model further enhances this performance, registering an EM of 50.66\% and an F1 score of 73.90\%, with notably reduced standard deviations, indicating consistent performance.

The German FinBERT model, when trained from scratch, exhibits an EM of 50.23\% and an F1 score of 72.80\%, which is comparable to the benchmark models. However, its further pre-trained iteration showcases an improvement to the benchmarks, achieving the highest EM of 52.50\% and F1 score of 74.61\% among all models.
\subsection{Translated Financial Phrase Bank (Finance-Specific Sentiment Classification Task)}
\begin{table}
\begin{flushleft}
\caption{Performance on Translated Financial Phrase Bank (Finance-Specific Sentiment Classification)}
 \mytablefont{This table compares the performance on the test split of translated financial phrase bank of the German FinBERT model, trained from scratch (SC) and further pre-trained (FP), with the following benchmarks: \textit{bert-base-german-cased}, \textit{gbert-base} \citep{gbert} \& \textit{gottbert-base} \citep{gottbert}. Every model is trained for five different seeds. I report the accuracy and F1 scores averaged among seeds for every model, together with the standard deviation. Numbers are given in percent.}
\label{paper3_tab_model_performance_financial_phrase_bank}
\mytablefont
\begin{tabularx}{\textwidth}{c *{3}{Y}}
\toprule
Model & Accuracy & F1 (Macro) & F1 (Micro) \\
\midrule
bert-base-german-cased & \shortstack[c]{95.03 \\ (0.29)} & \shortstack[c]{90.21 \\ (0.24)} & \shortstack[c]{92.54 \\ (0.43)} \\[0.3cm]
gbert-base & \shortstack[c]{94.61 \\ (0.24)} & \shortstack[c]{89.90 \\ (0.52)} & \shortstack[c]{91.91 \\ (0.35)} \\[0.3cm]
gottbert-base & \shortstack[c]{94.99 \\ (0.56)} & \shortstack[c]{90.11 \\ (1.24)} & \shortstack[c]{92.49 \\ (0.84)} \\
\midrule 
GermanFinBERT\textsubscript{SC} & \shortstack[c]{\textbf{95.95} \\ (0.24)} & \shortstack[c]{\textbf{92.70} \\ (0.62)} & \shortstack[c]{\textbf{93.93} \\ (0.35)} \\[0.3cm]
GermanFinBERT\textsubscript{FP} & \shortstack[c]{95.41 \\ (0.39)} & \shortstack[c]{91.49 \\ (0.83)} & \shortstack[c]{93.12 \\ (0.59)} \\
\bottomrule
\end{tabularx}
\end{flushleft}
\end{table}

Table \ref{paper3_tab_model_performance_financial_phrase_bank} shows the results for the financial phrase bank task \citep{malo2014good}, which was translated to German. As this is a sentiment classification task with three classes, I report the classification accuracy in conjunction with both micro and macro F1 scores. 
Among the benchmark models, the \textit{bert-base-german-cased} model performs best, registering an accuracy of 95.03\%, a macro F1 score of 90.21\% and a micro F1 score of 92.54\%. The remaining benchmark models exhibit comparable performance metrics, albeit marginally lower.

Notably, the German FinBERT model pre-trained from scratch surpasses all benchmark models, achieving an accuracy of 95.95\%, a macro F1 score of 92.70\% and a micro F1 score of 93.93\%. This is an improvement of 1-2 percentage points for all metrics compared to the benchmark models. The consistency of these results is underscored by a small standard deviation across all metrics. While the further pre-trained version of the German FinBERT model also exceeds the performance of all benchmark models, it marginally lags behind its counterpart trained from scratch with respect to all performance measures and their volatility.

Subsequently, I evaluate the performance of the models on the general downstream tasks to ascertain if the finance-specific FinBERT models also perform reliability in a more generic context.
\subsection{10k German News Articles Data Set}
\begin{table}
\begin{flushleft}
\caption{Performance on 10k GNAD database}
 \mytablefont{This table compares the performance on the test split of 10k GNAD database of the German FinBERT model, trained from scratch (SC) and further pre-trained (FP), with the following benchmarks: \textit{bert-base-german-cased}, \textit{gbert-base} \citep{gbert} \& \textit{gottbert-base} \citep{gottbert}. Every model is trained for five different seeds. I report the accuracy and F1 scores averaged among seeds for every model, together with the standard deviation. Numbers are given in percent.}
\label{paper3_tab_model_performance_10kGNAD}
\mytablefont
\begin{tabularx}{\textwidth}{c *{3}{Y}}
\toprule
Model & Accuracy & F1 (Macro) & F1 (Micro) \\
\midrule
bert-base-german-cased & \shortstack[c]{97.91 \\ (0.08)} & \shortstack[c]{90.16 \\ (0.45)} & \shortstack[c]{90.60 \\ (0.37)} \\[0.3cm]
gbert-base & \shortstack[c]{\textbf{98.06} \\ (0.08)} & \shortstack[c]{\textbf{91.09} \\ (0.42)} & \shortstack[c]{\textbf{91.28} \\ (0.37)} \\[0.3cm]
gottbert-base & \shortstack[c]{97.83 \\ (0.14)} & \shortstack[c]{90.01 \\ (0.71)} & \shortstack[c]{90.21 \\ (0.61)} \\
\midrule 
GermanFinBERT\textsubscript{SC} & \shortstack[c]{97.55 \\ (0.12)} & \shortstack[c]{88.42 \\ (0.48)} & \shortstack[c]{88.97 \\ (0.55)} \\[0.3cm]
GermanFinBERT\textsubscript{FP} & \shortstack[c]{98.01 \\ (0.09)} & \shortstack[c]{90.65 \\ (0.32)} & \shortstack[c]{91.05 \\ (0.41)} \\
\bottomrule
\end{tabularx}
\end{flushleft}
\end{table}

In the context of evaluating models on generic downstream tasks, the 10kGNAD task emerges as a suitable multi-class topic classification challenge. Table \ref{paper3_tab_model_performance_10kGNAD} shows the accuracy as well as the macro and micro F1 scores of all models.

The \textit{gbert-base} model performs best among all models, with an accuracy of 98.06\%, a macro F1 score of 91.09\% and a micro F1 score of 91.28\%. The \textit{bert-base-german-cased} model, while closely trailing with an accuracy of 97.91\%, achieves a macro F1 score of 90.16\% and a micro F1 score of 90.60\%. The \textit{gottbert-base} model exhibits slightly lower metrics, with an accuracy of 97.83\%, a macro F1 score of 90.01\% and a micro F1 score of 90.21\%.

The further pre-trained version of German FinBERT, while not outperforming the \textit{gbert-base}, still showcases comparable results. It achieves an accuracy of 98.01\%, a macro F1 score of 90.65\% and a micro F1 score of 91.05\%. When trained from scratch, the German FinBERT exhibits a slightly lower performance compared to the other models, though the difference in performance is small.

\subsection{German QuAD}
\begin{table}
\begin{flushleft}
\caption{Performance on GermanQuAD}
 \mytablefont{This table compares the performance on the test split of GermanQuAD data set of the German FinBERT model, trained from scratch (SC) and further pre-trained (FP), with the following benchmarks: \textit{bert-base-german-cased}, \textit{gbert-base} \citep{gbert} \& \textit{gottbert-base} \citep{gottbert}. Every model is trained for five different seeds. I report the exact matches (EM) and F1 scores \citep{moller2021germanquad} averaged among seeds for every model, together with the standard deviation. Numbers are given in percent.}
\label{paper3_tab_model_performance_german_quad}
\mytablefont
\begin{tabularx}{\textwidth}{c *{2}{Y}}
\toprule
Model & EM & F1 \\
\midrule
bert-base-german-cased & \shortstack[c]{48.65 \\ (0.47)} & \shortstack[c]{67.83 \\ (0.24)} \\[0.3cm]
gbert-base & \shortstack[c]{54.98 \\ (0.80)} & \shortstack[c]{72.82 \\ (0.83)} \\[0.3cm]
gottbert-base & \shortstack[c]{\textbf{55.14} \\ (0.50)} & \shortstack[c]{\textbf{73.06} \\ (0.48)} \\
\midrule 
GermanFinBERT\textsubscript{SC} & \shortstack[c]{47.96 \\ (0.11)} & \shortstack[c]{64.92 \\ (0.29)} \\[0.3cm]
GermanFinBERT\textsubscript{FP} & \shortstack[c]{54.26 \\ (0.39)} & \shortstack[c]{72.28 \\ (0.49)} \\
\bottomrule
\end{tabularx}
\end{flushleft}
\end{table}

Table \ref{paper3_tab_model_performance_german_quad} shows the results for the German QuAD task, with the same performance measures that I use for the finance-specific ad-hoc QuAD task. 

The \textit{gottbert-base} model emerges as best performing model, achieving an EM score of 55.14\% and an F1 score of 73.06\%. Both these metrics are accompanied by relatively low standard deviations, indicating consistent performance across different training seeds. The \textit{gbert-base} model closely follows, with only marginal differences in the scores. In contrast, the \textit{bert-base-german-cased} model lags slightly behind, particularly in the F1 score.

The German FinBERT model, when trained from scratch, exhibits a comparable performance as the \textit{bert-base-german-cased} model, with an EM of 47.96\% and an F1 score of 64.92\%. However, its further pre-trained counterpart demonstrates a significant improvement, nearly matching the performance of the \textit{gbert-base} model.

\subsection{GermEval 2018 (Generic Sentiment Classification Task)}
\begin{table}
\begin{flushleft}
\caption{Performance on GermEval 2018 database (Generic Sentiment Classification)}
 \mytablefont{This table compares the performance on the test split of GermEval 2018 database of the German FinBERT model, trained from scratch (SC) and further pre-trained (FP), with the following benchmarks: \textit{bert-base-german-cased}, \textit{gbert-base} \citep{gbert} \& \textit{gottbert-base} \citep{gottbert}. Every model is trained for five different seeds. I report the accuracy and macro F1 score averaged among seeds for every model, together with the standard deviation. Numbers are given in percent.}
\label{paper3_tab_model_performance_germ_eval_2018}
\mytablefont
\begin{tabularx}{\textwidth}{c *{2}{Y}}
\toprule
Model & Accuracy & F1 (Macro) \\
\midrule
bert-base-german-cased & \shortstack[c]{78.84 \\ (0.18)} & \shortstack[c]{74.56 \\ (0.22)} \\[0.3cm]
gbert-base & \shortstack[c]{80.91 \\ (0.45)} & \shortstack[c]{76.86 \\ (0.59)} \\[0.3cm]
gottbert-base & \shortstack[c]{\textbf{81.25} \\ (0.14)} & \shortstack[c]{\textbf{77.86} \\ (0.32)} \\
\midrule 
GermanFinBERT\textsubscript{SC} & \shortstack[c]{77.14 \\ (0.65)} & \shortstack[c]{72.75 \\ (0.98)} \\[0.3cm]
GermanFinBERT\textsubscript{FP} & \shortstack[c]{80.75 \\ (0.70)} & \shortstack[c]{76.70 \\ (1.19)} \\
\bottomrule
\end{tabularx}
\end{flushleft}
\end{table}

In the concluding generic downstream evaluation, we assess the models on the sentiment classification task derived from the GermEval 2018 challenge \citep{risch2018fine}. Table \ref{paper3_tab_model_performance_germ_eval_2018} delineates the accuracy and macro F1 score for each model under consideration. Note that, unlike the finance-specific sentiment classification, the micro F1 score is omitted in this context due to the binary nature of the classification, resulting in a congruence between accuracy and the micro F1 score.

The \textit{gottbert-base} model achieves the highest performance across all models, registering an accuracy of 81.25\% and a macro F1 score of 77.86\%. This is closely followed by the \textit{gbert-base} model, which records an accuracy of 80.91\% and a macro F1 score of 76.86\%. The \textit{bert-base-german-cased} model, while competitive, lags slightly behind with an accuracy of 78.84\% and a macro F1 score of 74.56\%.

Turning our attention to the German FinBERT model, the further pre-trained version (FP) demonstrates assimilable results with an accuracy of 80.75\% and a macro F1 score of 76.70\%. This is notably higher than its counterpart trained from scratch, which posts an accuracy of 77.14\% and a macro F1 score of 72.75\%.

\subsection{Zero-Shot Passage Retrieval by Topic}

The previous results predominantly centered on the models' performance post fine-tuning on specific downstream tasks. However, an exploration into the German FinBERT's zero-shot capabilities on a finance-specific task offers an alternative perspective. Such an evaluation can potentially shed light on the model's proficiency in grasping and representing the nuanced intricacies inherent to financial textual data after its pre-training phase. To this end, I compute the zero-shot passage retrieval performance of both German FinBERT variants and all benchmarks on a finance-specific data set.

The essence of the used passage retrieval task lies in extracting passages from a corpus of ad-hoc announcements that coincide with the topic of a given query. To achieve this, I first manually create 10 to 20 queries for each of the 20 topics from the ad-hoc multi-label database, resulting in 251 queries in total. Subsequently, I transform the ad-hoc multi-label database to a collection of paragraphs, each paragraph composed of 2 or 3 sentences, with aggregated labels. For each topic, I sample 500 paragraphs from the entire corpus to achieve a more balanced distribution across topics. The next step involves embedding both the queries and the paragraphs using the [CLS] token's embedding from the BERT models. The final stage of this process is the computation of pairwise cosine similarities between these embeddings.

The underlying hypothesis driving this retrieval process is that paragraphs exhibiting the highest similarity to a given query are likely to be topically aligned with that query.

\begin{figure}[h]
    \centering
    \includegraphics[width=0.8\textwidth]{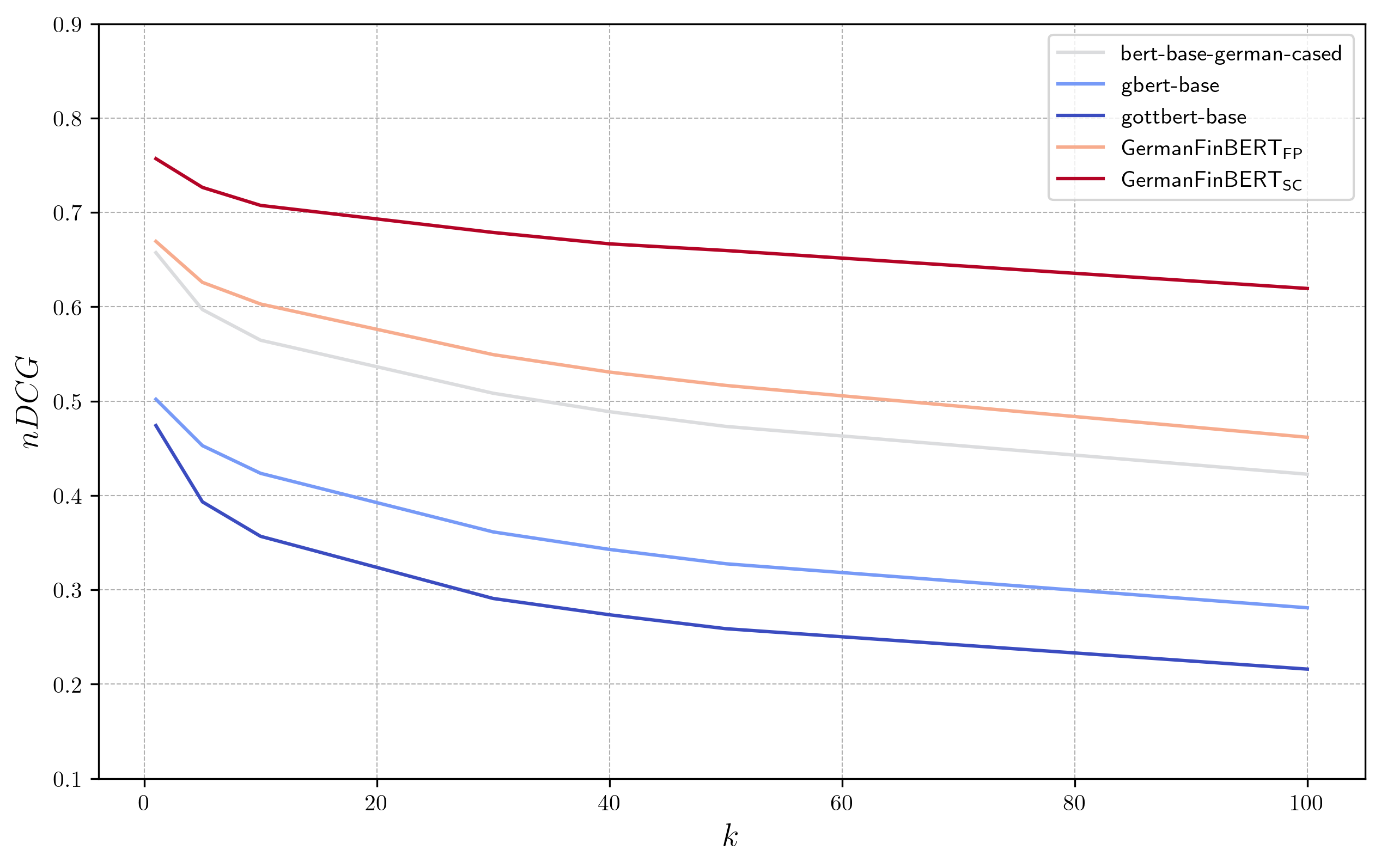}
    \caption{\textit{nDCG} Performance for Zero-Shot Ad-Hoc Passage Retrieval by Topic}
    \label{paper_3_ndcg_performance}
\end{figure}

To quantify the effectiveness of this retrieval, I employ the normalized discounted cumulative gain (\textit{nDCG}) of \cite{jarvelin2002cumulated}. \textit{nDCG} is a measure that evaluates the quality of a set of search results by considering both the relevance of retrieved documents and their rank position. Given the nature of this task, where the relevance of results diminishes as one moves down the list of results, \textit{nDCG} emerges as a fitting metric. For the evaluation, the relevance scores are kept binary: a score of 1 indicates the presence of the query topic in the retrieved paragraph, while a score of 0 denotes its absence. The computation of the \textit{nDCG} measure can be reduced to the top k search results. Figure \ref{paper_3_ndcg_performance} shows the passage retrieval performances, measured by \textit{nDCG}, for all models and for values of k between 1 and 100. 

In the evaluation, both variants of the German FinBERT model consistently surpass the benchmark models across all considered values of k. Moreover, pre-training the German FinBERT model from scratch shows markedly greater efficacy than the further pre-training approach for the considered task. This distinction is evident in the discernible gap between the \textit{nDCG} performance curves of the two German FinBERT variants, as depicted in Figure \ref{paper_3_ndcg_performance}, with a disparity exceeding 10 percentage points for higher values of k. This suggests that while both methodologies enhance the model's capability to encode finance-specific information, there is evidence that pre-training from scratch is more proficient in this context.

\subsection{Discussion}
The results of the downstream tasks yield several insights. With respect to the finance specific downstream tasks, we see that the further pre-trained version of the German FinBERT performs best among all models across for the ad-hoc multi-label and QuAD tasks, while the pre-trained version from scratch outperforms all other models for the sentiment classification task. These results underscore the potential advantages of additional pre-training of BERT models on domain specific data. Based on the findings of this study, pre-training from scratch does not appear to offer a distinct advantage, even if it performs best for sentiment classification, as pre-training from scratch is more time-consuming, data-intensive and expensive. 

Upon examining the generic downstream tasks, we observe that the further pre-trained variant of the German FinBERT aligns closely with the performance of benchmark models, albeit with a slightly declined performance. As the \textit{gbert-base} model serves as the starting point for the further pre-trained German FinBERT, a comprehensive comparison between their performances is meaningful. Notably, the further pre-trained German FinBERT surpasses the \textit{gbert-base} in all finance-specific downstream tasks. Conversely, for generic downstream tasks, the \textit{gbert-base} takes precedence. This suggests a potential trade-off: while further pre-training augments the model's efficacy in domain-specific tasks, it might marginally compromise its prowess in generic tasks. Nonetheless, the disparities in performance are small, implying that both models maintain sufficient proficiency across tasks. The performance of the German FinBERT model, when pre-trained from scratch, on generic downstream tasks is consistently inferior to that of benchmark models. This suggests a decline in model quality when the textual data the model is applied on is not aligned with the financial domain.

When juxtaposed with existing literature on domain-specific pre-training of BERT models, the solid results from the further pre-trained version of the German FinBERT align with prior research findings. Contrarily, the mixed outcomes from the German FinBERT trained from scratch are unexpected, given its superior performance as documented in earlier studies. One  explanation for this discrepancy could be the model's possible undertraining. Notably, I observed a constant increase in downstream performance for finance-related tasks upon the addition of further pre-training steps. The German FinBERT model trained from scratch, as discussed in this paper, underwent pre-training for 174,000 steps, whereas the original BERT model from \cite{bert} was subjected to 1,000,000 steps. The constraints on further pre-training steps were due to limited three-day GPU access, followed by requisite re-registration and associated queuing delays. Nevertheless, I will continue the pre-training of the model from scratch until there's no further improvement in performance on the downstream tasks.\\
Another potential factor influencing the performance of the German FinBERT model trained from scratch could be the high share of annual reports within the pre-training corpus. These reports often encompass standardized sections that recur across various companies, suggesting that despite the corpus's volume, there might be limited content variance. Future studies could consider enriching the corpus with a broader array of data or refining it through the elimination of duplicates.

Among the benchmark models, no single model consistently surpassed the others. Each model demonstrates superiority in distinct downstream tasks, suggesting that their performances are mostly similar across all tasks. 

\section{Conclusion}
In the rapidly evolving digital landscape, the ability to extract meaningful insights from vast amounts of unstructured textual data has become paramount, especially in the financial domain. This research has underscored the significance of domain-specific pre-training in enhancing the performance of language models, as evidenced by the development and evaluation of the German FinBERT model tailored for financial texts.

Drawing from a rich data set, primarily sourced from the German Federal Gazette (Bundesanzeiger) and supplemented with news articles, ad-hoc announcements and Wikipedia articles, the German FinBERT has been trained to capture the unique lexicon and semantic structures inherent to German financial literature. The exploration of two distinct training methodologies, training from scratch and further pre-training of a generic model, has provided valuable insights into the nuances of domain-specific pre-training.

The superior performance of the German FinBERT, when compared to generic German BERT models on finance-specific downstream tasks, reaffirms the hypothesis that domain-specific models offer tangible advantages in specialized sectors. This research does not only fill a notable gap in the realm of domain-specific models for the German financial market but also sets a precedent for similar endeavors in other languages and sectors.

As the financial world continues to integrate technology and data-driven decision-making, tools like the German FinBERT will play an instrumental role in harnessing the potential of unstructured data. Future research can build upon this foundation, exploring further refinements in training methodologies, expanding the data set and adapting the model for even more specific financial niches.

\newpage
\bibliographystyle{jf}
\bibliography{References.bib}
\newpage

\section*{Appendix}
\addcontentsline{toc}{section}{Appendix}

\subsection*{Details on the Generation of the Ad-Hoc QuAD Database}
To construct the ad-hoc QuAD database, I use all ad-hoc announcements from the multi-label database presented in \cite{AdHocMultilabel}. Announcements exceeding 15 sentences are truncated to ensure compatibility with BERT's input limitations in subsequent applications.

Announcements from the ad-hoc multi-label database are utilized as context strings. Subsequently, there is a need to identify questions and appropriate answers that reference these context strings. Given that manual generation of questions and answers is both resource-intensive and time-consuming, I employ the OpenAI's ChatGPT model \citep{ChatGPT} for this purpose. Specifically, the \textit{gpt-3.5-turbo} version of ChatGPT is chosen with its default settings, as it presents an optimal balance between performance, speed and costs.

In a first step, I ask ChatGPT to generate three suitable questions for a given announcement. The prompt looks as follows:
\begin{displayquote}
\begin{ttfamily}
Create three questions for the following text. It should be possible to answer the question with a substring of the input text. The questions should ask for different aspects of the input. The questions should be in German.\\ \\Text: <<context>>\\Question:
\end{ttfamily}
\end{displayquote}
In the pursuit of creating an extractive QuAD task, it is imperative to instruct the model such that every question can be answered using a substring from the provided announcement. This strategy aims to prevent the model from generating open-ended questions or those requiring external knowledge not present in the announcement. Additionally, the model is directed to address various aspects of the announcement to minimize question redundancy. Notably, despite the context strings being in German, ChatGPT occasionally formulates questions in English. To counteract this, explicit instructions are given to ensure questions are posed in German. Employing this methodology yields 9,132 unique context-question pairs.
\newpage
In a second step, I use ChatGPT again to extract the substring that answers to question to a specific context string. The respective prompt is given by:
\begin{displayquote}
\begin{ttfamily}
You have given a text and a question to that text. Find the answer as a substring of the input text. It is crucial that the answer is contained exactly as a substring in the input text, even if this implies that the answer is not a full sentence. Example: \\\\Text: 'Herr M\"uller ist 37 Jahre alt.'\\Question: 'Wie alt ist Herr M\"uller?'\\ Answer: '37 Jahre'\\ \\Text: <<context>>\\Question: <<question>>\\Answer:
\end{ttfamily}
\end{displayquote}
Evaluations of the method of extracting substrings from a specified context to answer a posed question via ChatGPT indicated a recurrent issue: ChatGPT frequently transformed the substring into a complete sentence, thereby compromising the extractive nature of the resultant database. Emphasizing the necessity for extractive answers, coupled with a demonstrative example, markedly enhanced the outcomes. However, of the responses generated by ChatGPT, 1,725 are not given as substrings of the context, leading to a final ad-hoc QuAD database size of 7,407.

\end{document}